\documentclass{article}

\usepackage{arxiv}
\usepackage{multirow} 
\usepackage[utf8]{inputenc} 
\usepackage[T1]{fontenc}    
\usepackage{hyperref}       
\usepackage{url}            
\usepackage{booktabs}       
\usepackage{amsfonts}       
\usepackage{nicefrac}       
\usepackage{microtype}      
\usepackage{lipsum}		
\usepackage{graphicx}
\usepackage[numbers]{natbib}
\usepackage{doi}
\usepackage{amsmath}

\title{ICPC: In-context Prompt Compression with Faster Inference}

\author{ Ziyang Yu\\
	Department of Mathematics\\
	Southern University of Science and Technology\\
	Shenzhen, China  \\
	\texttt{11910419@mail.sustech.edu.cn} \\
	\And
	Yuyu Liu\\
	Department of Mathematics\\
	Southern University of Science and Technology\\
	Shenzhen, China  \\
	\texttt{liuyy2020@mail.sustech.edu.cn} \\
}

\date{}

\hypersetup{
pdftitle={A template for the arxiv style},
pdfsubject={q-bio.NC, q-bio.QM},
pdfauthor={David S.~Hippocampus, Elias D.~Striatum},
pdfkeywords={First keyword, Second keyword, More},
}

\begin{document}
\maketitle

\begin{abstract}
Despite the recent success of Large Language Models
(LLMs), it remains challenging to feed LLMs with long prompts due to the fixed size of LLM inputs. As a remedy, prompt compression becomes a promising solution by removing redundant tokens in the prompt. However, using LLM in the existing works requires additional computation resources and leads to memory overheads. To address it, we propose ICPC (In-context Prompt Compression), a novel
and scalable prompt compression method that adaptively reduces the prompt length. The key idea of ICPC is to calculate the probability of each word appearing in the prompt using encoders and calculate information carried by each word through the information function, which effectively
reduces the information loss during prompt compression and increases the speed of compression. Empirically, we demonstrate that ICPC can
effectively compress long texts of different categories and thus achieve better
performance and speed on different types of NLP tasks.
\end{abstract}


\section{Introduction}

\emph{Large Language Models} (LLMs) have revolutionized natural language processing, demonstrating remarkable capabilities across various tasks such as text generation, question answering, and semantic understanding~\cite{zhao2023survey, minaee2024large, hadi2023survey}. However, LLMs encounter significant challenges when processing long prompts or extended contexts, as their computational cost scales quadratically with sequence length due to the attention mechanism~\cite{li2024prompt, zhu2024survey, bai2024beyond}. This limitation hinders their ability to handle real-world applications requiring long-context understanding, such as document summarization, multi-turn conversations, and knowledge-intensive reasoning. Researchers have explored efficient attention mechanisms, sparse representations, and distributed processing approaches to address these scalability issues, enabling LLMs to manage long contexts more effectively and maintain their performance on extended inputs. However, existing approaches often utilize LLM to compress texts, leading to memory overhead challenges.

There are numerous existing approaches aimed at addressing the memory overhead of large language models (LLMs) during inference, including LoRA~\cite{hu2021lora} and Sparse Attention~\cite{child2019generating}. Several prompt compression methods, such as Selective Context~\cite{li2023compressing}, have demonstrated strong performance with limited memory requirements. Unlike these existing methods, which primarily focus on utilizing large language models to compress prompts, we propose a novel approach to prompt compression leveraging language models with millions of parameters. Our approach offers faster inference compared to existing methods and achieves excellent compression performance.

The motivation for the proposed approach stems from the structure of transformer encoders. First, the pretraining of transformer encoders enables them to capture and understand the context of words effectively. Second, transformer encoders typically have significantly fewer parameters than large language models, resulting in a 10x to 100x increase in inference speed over existing compression methods, shown in Appendix~\ref{tab:bert_vs_gpt3}.

In this paper, we introduce a novel framework named \underline{\textbf{I}}n-\underline{\textbf{C}}ontext \underline{\textbf{P}}rompt \underline{\textbf{C}}ompression (ICPC). ICPC operates by first segmenting sentences or paragraphs at both the phrase and clause levels. It then uses a pre-trained transformer encoder to calculate the loss associated with each word in the sentence, removing words to make the paragraph concise without loss of essential meaning. We conducted experiments across various encoder architectures to demonstrate the generalizability and superiority of our method over existing approaches.

\section{Preliminaries}

\subsection{Entropy}

Entropy, rooted in information theory, measures the average level of uncertainty or surprise in a probability distribution over linguistic units~\cite{shannon1948mathematical}. In Natural Language Processing (NLP), each token \(t \in T\) (e.g., word or subword) is associated with a probability \(p(t)\) reflecting how frequently \(t\) appears in a given context. Formally, the Shannon entropy of \(p\) is defined as

\[
H(p) = -\sum_{t \in T} p(t) \log p(t),
\]

where \(p(t)\) captures the likelihood of observing token \(t\). Entropy thus provides a principled way of quantifying how “spread out” or “concentrated” the distribution is. In language modeling, for example, a lower entropy typically indicates a model that makes confident predictions about the next token, whereas a higher entropy signals more uncertainty. Extensions like cross-entropy and KL divergence further leverage this concept to compare model-generated distributions against ground-truth or target distributions. Minimizing these metrics during training encourages NLP models to produce more accurate and reliable predictions, ultimately improving language understanding and generation.

\subsection{Masked Language Modeling}

The Transformer encoder, as popularized by models like BERT, leverages self-attention to capture contextual dependencies between tokens in a sentence. Instead of processing input sequentially, it computes pairwise interactions between tokens in parallel, enabling more efficient handling of long-range dependencies. A key training objective for Transformer encoders is Masked Language Modeling (MLM). In MLM, a subset of the input tokens is randomly masked (e.g., replaced with a special `[MASK]` token), and the model learns to predict these masked tokens based on their surrounding context. Formally, if \(x_{i}\) represents a masked token, the model estimates \(p(x_{i} \mid x_{\setminus i})\). By inferring missing pieces of text, the model learns robust internal representations that benefit downstream tasks such as text classification, question answering, and semantic similarity.
\section{Method}

Our method compresses the input text for LLM by removing redundant words and phrases to increase the ability of LLM understanding on long context understanding and reduce the computational cost without extra large language models (e.g., GPT-4, Llama-3) needed~\cite{achiam2023gpt, touvron2023llama}.

\subsection{Participle}

If participle-based filtering is directly applied at the word level, it may fail to capture the nuanced structure of linguistic patterns. Therefore, we implement participle-based filtering beyond word-level processing at both phrase and clause levels. In our framework, a participle unit is a fundamental building block, encompassing words, phrases, or clauses depending on the required granularity. To support this, we group tokens with contextual embeddings into participle units, enabling the model to retain richer semantic and syntactic information during filtering~\cite{li2023compressing}.

\subsection{Loss Computation}

Given a list of words $C=(x_{i-k},...,x_{i+k})$, the loss function by removing $x_i$ is defined as 
\begin{equation}
L(x_i) = \alpha * \sum_{\substack{n=-k \\ n \neq 0}}^{k} sim(\mathbf{x}_{i+n}, \mathbf{x}_{i}) + \log p(x_i \mid \mathbf{x}_{i, k})
\end{equation}
where $\mathbf{x}_{i, k}$ is defined as 
\begin{equation}
(x_{i-k}, \dots, x_{i-1}, x_{i+1}, \dots, x_{i+k} )
\end{equation}

This loss function provides a mechanism to balance the trade-off between compression and the preservation of information.



\subsection{Redundant Words Removal}
To minimize the loss of information while retaining the original key information during compression, we rank the units based on their calculated loss in descending order and compute the p-th percentile of loss among all units.
\begin{equation}
    L_p = np.percentile([L(x_0), .., L(x_n)], p)
\end{equation}

Then, we remove all the lexical units that will lose greater or equal to $L_p$ and merge the remaining words as the output $C^{'}$:

\begin{equation}
    C^{'} = \{x_i \mid L(x_i) < L_p\}
\end{equation}

This adaptive filtering strategy provides a more flexible mechanism to discard redundant units while retaining the most essential content. By dynamically adjusting the threshold, the method ensures that the selection process accounts for variations in the loss distribution.

\section{Experiments}
\label{sec:Experiments}
In this section, we present the performance of our method against other state-of-the-art approaches with both quantitative and qualitative analysis. For all experiments, we simulate an evaluation environment using an EC2 {\texttt{p4d.24xlarge}} virtual machine (VM) instance on AWS, which has $8$ NVIDIA A100 GPUs, $96$ vCPUs, and $320$~GB main memory. Other important
information including operation system version, Linux
kernel version and CUDA version are summarized in
Table~\ref{tab:versions}.

\subsection{Experimental Settings}

For a fair comparison, we adopt the same input format (tokens, phrases, or sentences) and inference settings for all experimental conditions. For parameters specific to our method, such as the compression ratio and lexical unit granularity, we tune these to optimize efficiency without degrading performance. Baseline methods such as random compression and full context usage retain default configurations. All metrics (e.g., BLEU) are computed under identical evaluation protocols to ensure consistency across tasks, with multiple runs for stochastic outputs to mitigate randomness.

\subsubsection{Datasets}

Our method reduces redundancy in the input context, enabling efficient processing of very long contexts for LLMs. However, existing benchmarks such as SQuAD~\cite{rajpurkar2018know} and Piqa~\cite{bisk2020piqa} are mostly single-round question-answer datasets with short question length, which is not appropriate to evaluate our proposed method. Therefore, we compile four datasets with long context and conversations to demonstrate the effectiveness of our method. The statistics and compilation details are presented in the appendix.

\noindent \textbf{Wikipedia}: A dataset containing articles from Wikipedia, a free online encyclopedia covering an extensive range of topics across numerous domains, including history, science, arts, and culture. For our experiments, we utilize the introductory sections of each article, which provide concise and informative summaries of the topics.

\noindent \textbf{arXiv Papers}: A dataset comprising academic papers from arXiv, spanning diverse fields such as computer science, physics, mathematics, and biology. Due to the length of many arXiv papers, we focus on processing our experiments' abstract and introduction sections, ensuring a balance between content depth and computational efficiency.

\noindent \textbf{Reddit}: A dataset derived from user-generated posts and comments on Reddit, a social media platform organized into communities covering various topics, from technology and science to hobbies and entertainment. For our experiments, we use a curated subset of posts and their associated top-level comments to capture meaningful discussions and interactions.

\subsubsection{Models}
To demonstrate the generalization of our method in different settings, we test the method on different kinds of language models. We evaluate our method on BERT~\cite{devlin2018bert}, RoBERTa~\cite{liu2019roberta}, XLNet~\cite{yang2019xlnet}, ALBERT~\cite{lan2019albert}, T5~\cite{raffel2020exploring}, DeBERTa~\cite{he2020deberta}. Appendix~\ref{app:encoders} shows the detailed descriptions of encoders.

\subsubsection{Metrics}
We evaluate our method using BLEU~\cite{papineni2002bleu}, ROUGE~\cite{lin2004rouge}, TF-IDF Similarity~\cite{sparck1972statistical}, Jaccard Similarity~\cite{jaccard1912distribution}, BERTScore~\cite{zhang2019bertscore}, compression Rate~\cite{roberts2003benefits}, and Flesch-Kincaid readability score~\cite{kincaid1975derivation}. Appendix~\ref{app:metrics} shows the detailed description of these encoders.

\subsection{ICPC Performance Evaluation}

\begin{table*}[t!]
\scriptsize
\centering
\begin{tabular}{ccccccccccccccc}

\toprule
\multirow{3}{*}{\textbf{Model}} & \multirow{3}{*}{\textbf{Ratio}} & \multirow{3}{*}{\textbf{METEOR}} & \multirow{3}{*}{\textbf{BLEU}} & \multicolumn{3}{c}{\textbf{ROUGE}} & \multicolumn{3}{c}{\textbf{BERTScore}} \\ 
\cmidrule(lr){5-7} \cmidrule(lr){8-10}
& & & & \textbf{rouge1} & \textbf{rouge2} & \textbf{rougeL}
& \textbf{Precision} & \textbf{Recall} & \textbf{F1} \\

\midrule
Original & - & 49.3 & 44.1 & 66.9 & 49.8 & 56.6 &  86.7 & 90.9 & 88.8  \\
\midrule
\multirow{3}{*}{Random Deletion}  
& 0.8 & 43.9 & 40.8 & 61.9 & 43.1 & 52.1 & 81.9 & 81.3 & 81.5 \\
& 0.6 & 41.6 & 36.1 & 58.1 & 37.5 & 47.5 & 78.1 & 79.5 & 79.2 \\
& 0.4 & 39.2 & 31.8 & 53.9 & 32.4 & 43.2 & 75.0 & 75.1 & 75.4 \\
\midrule
\multirow{3}{*}{Selective Context}  
& 0.8 & 45.1 & 41.7 & 62.7 & 43.8 & 52.7 & 82.9 & 82.9 & 82.9 \\
& 0.6 & 42.6 & 37.2 & 58.8 & 38.2 & 48.2 & 79.2 & 80.4 & 79.8 \\
& 0.4 & 39.9 & 32.6 & 54.3 & 33.2 & 44.3 & 76.1 & 76.0 & 76.0 \\
\midrule
\multirow{3}{*}{LLMLingua}  
& 0.8 & 45.2 & 42.1 & 61.8 & 43.1 & 52.1 & 83.4 & 83.1 & 83.2 \\
& 0.6 & 43.1 & 37.8 & 58.4 & 37.9 & 47.9 & 78.0 & 79.1 & 78.5 \\
& 0.4 & 40.1 & 32.9 & 55.0 & 33.4 & 44.8 & 76.3 & 75.6 & 75.9 \\
\midrule
\multirow{3}{*}{\textbf{ICPC}}  
& 0.8 & 45.3 & 42.7 & 63.1 & 44.2 & 52.8 & 83.6 & 83.5 & 83.5 \\
& 0.6 & 43.4 & 38.0 & 59.1 & 38.4 & 48.5 & 79.1 & 80.6 & 79.8 \\
& 0.4 & 40.6 & 33.1 & 55.2 & 33.6 & 45.1 & 76.4 & 76.2 & 76.3 \\
\bottomrule
\end{tabular}%
\caption{Performance comparison of baseline methods. The ICPC utilize BERT as encoder. ICPC can boost
the performance compared to traditional methods using large language models.}
\label{tab: Performance comparison}
\vspace{-4mm}
\end{table*}

\begin{table*}[h]
\scriptsize
\centering
\begin{tabular}{ccccccccccccccc}

\toprule
\textbf{Model} & \textbf{Ratio} & \textbf{Training time (ms)} \\

\midrule
\multirow{3}{*}{Selective Context}  
& 0.8 & 46.3  \\
& 0.6 & 49.5  \\
& 0.4 & 52.2  \\
\midrule
\multirow{3}{*}{LLMLingua}  
& 0.8 & 45.2 \\
& 0.6 & 43.1 \\
& 0.4 & 40.1 \\
\midrule
\multirow{3}{*}{\textbf{ICPC}}  
& 0.8 & 10.3 \\
& 0.6 & 13.4 \\
& 0.4 & 16.6 \\
\bottomrule
\end{tabular}%
\caption{Average compression time comparison of baseline methods. The ICPC utilizes BERT as the encoder. ICPC can significantly reduce compression time compared to traditional methods.}
\label{tab: Compression time comparison}
\vspace{-4mm}
\end{table*}
In this section, we offer an in-depth evaluation of the performance enhancements attributed to our In-context Prompt Compression (ICPC). Specifically, we evaluate ICPC against 4 baseline methods, including Original, Random Deletion, Selective Context, and LLMLingua. These baseline methods do not require model training.

As shown in Table ~\ref{tab: Performance comparison}, it’s evident that the ICPC has led to marked improvements across all metrics. Notably, ICPC saw the BLEU score increase with ratio 0.6 on metric BLEU from 42.6 to 43.4, highlighting ICPC’s ability in understanding the in-context information of prompt. Random deletion underperforms
mainly due to its limited importance understanding of words in the prompt, which led developers to make selections on the prompt according to the importance score of different words.
LLMLingua performs well on different metrics
using token-level iterative compression and distribution alignment between models, though it faces slowdowns due to the usage of large language models.

\subsection{Compression Speed Analysis}

As shown in Table ~\ref{tab: Compression time comparison}, the ICPC achieves faster compression speed compared to Selective Context and LLMLingua. It is evident that the ICPC reduces compression time multiple times by using models of smaller size. It is noted that the performance does not degrade as our method utilizes context information instead of previous information, such as GPT and Llama, etc. 

\subsection{Readability Analysis}

As shown in Figure 2, the ICPC compresses the prompt by calculating the importance of each word in the sentence and removing unnecessary words to make the prompts shorter and more concise without losing information. The compressed text also preserves good readability and makes it easier for people to grasp the meaning of the long prompt.

\begin{figure*}[h!]
\centering
\includegraphics[width=\textwidth]{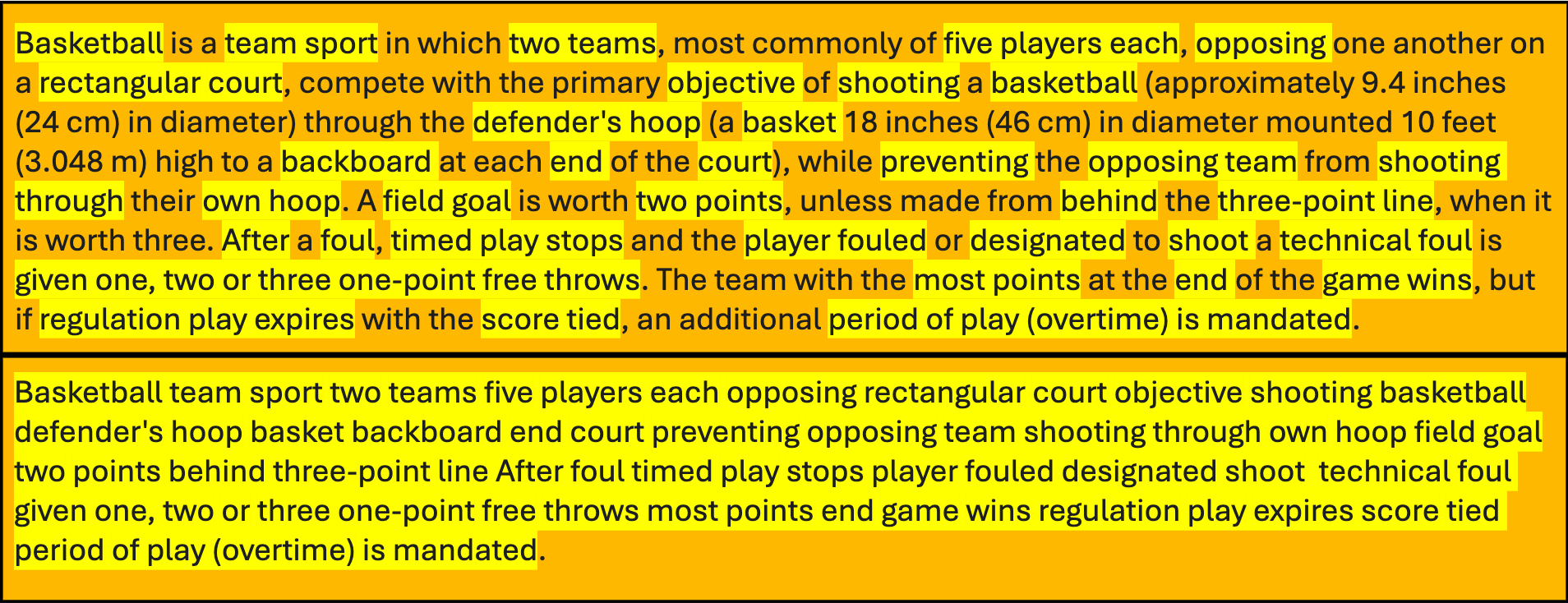}
\caption{Texts before and after compression. Yellow represents words with higher importance. Up: text before compression. Down: text after compression.}
\label{Texts before and after compression}
\end{figure*}

\begin{table*}[h!]
\scriptsize
\centering
\begin{tabular}{ccccccccccccccc}

\toprule
\multirow{3}{*}{\textbf{Model}} & \multirow{3}{*}{\textbf{Ratio}} & \multirow{3}{*}{\textbf{METEOR}} & \multirow{3}{*}{\textbf{BLEU}} & \multicolumn{3}{c}{\textbf{ROUGE}} & \multicolumn{3}{c}{\textbf{BERTScore}} \\ 
\cmidrule(lr){5-7} \cmidrule(lr){8-10}
& & & & \textbf{rouge1} & \textbf{rouge2} & \textbf{rougeL}
& \textbf{Precision} & \textbf{Recall} & \textbf{F1} \\

\midrule
Original & - & 49.3 & 44.1 & 66.9 & 49.8 & 56.6 &  86.7 & 90.9 & 88.8  \\
\midrule
\multirow{3}{*}{\textbf{BERT}}  
& 0.8 & 45.3 & 42.7 & 63.1 & 44.2 & 52.8 & 83.6 & 83.5 & 83.5 \\
& 0.6 & 43.4 & 38.0 & 59.1 & 38.4 & 48.5 & 79.1 & 80.6 & 79.8 \\
& 0.4 & 40.6 & 33.1 & 55.2 & 33.6 & 45.1 & 76.4 & 76.2 & 76.3 \\
\midrule
\multirow{3}{*}{\textbf{RoBERTa}}  
& 0.8 & 45.1 & 42.2 & 63.0 & 44.5 & 52.4 & 83.3 & 83.6 & 83.4 \\
& 0.6 & 43.3 & 38.1 & 59.3 & 38.6 & 48.5 & 79.6 & 80.4 & 80.0 \\
& 0.4 & 40.5 & 33.6 & 55.1 & 33.3 & 45.3 & 76.2 & 76.3 & 76.2 \\
\midrule
\multirow{3}{*}{\textbf{XLNet}}  
& 0.8 & 45.5 & 42.4 & 63.1 & 44.3 & 52.1 & 83.6 & 83.2 & 83.4 \\
& 0.6 & 43.3 & 38.4 & 59.4 & 38.2 & 48.2 & 79.7 & 80.2 & 79.9 \\
& 0.4 & 40.7 & 33.2 & 55.7 & 33.5 & 45.5 & 76.4 & 76.9 & 76.6 \\
\midrule
\multirow{3}{*}{\textbf{ALBERT}}  
& 0.8 & 44.9 & 42.4 & 62.8 & 44.1 & 52.4 & 83.2 & 83.2 & 83.2 \\
& 0.6 & 42.9 & 38.5 & 59.5 & 38.6 & 48.1 & 79.5 & 80.3 & 79.9 \\
& 0.4 & 40.8 & 33.2 & 55.6 & 33.3 & 45.6 & 76.9 & 76.1 & 76.5 \\
\midrule
\multirow{3}{*}{\textbf{T5}}  
& 0.8 & 45.5 & 42.2 & 61.9 & 43.5 & 52.3 & 83.8 & 83.5 & 83.4 \\
& 0.6 & 43.2 & 37.3 & 58.2 & 38.0 & 47.2 & 78.4 & 79.2 & 78.8 \\
& 0.4 & 40.6 & 32.5 & 55.9 & 33.8 & 44.9 & 76.2 & 75.7 & 75.9 \\
\midrule
\multirow{3}{*}{\textbf{DeBERTa}}  
& 0.8 & 45.9 & 42.9 & 61.2 & 43.8 & 52.7 & 83.3 & 83.6 & 83.4 \\
& 0.6 & 43.4 & 37.5 & 58.6 & 38.9 & 47.3 & 78.5 & 79.2 & 78.8 \\
& 0.4 & 40.2 & 32.3 & 55.7 & 33.6 & 44.6 & 76.1 & 75.5 & 75.8 \\
\bottomrule
\end{tabular}%
\caption{Performance comparison of ICPC using different encoders. ICPC utilizes 6 different types of models as encoder: BERT, RoBERTa, XLNet, ALBERT, T5 and DeBERTa.}
\label{tab: different encoders}
\vspace{-4mm}
\end{table*}

\begin{table*}[h!]
\scriptsize
\centering
\begin{tabular}{@{}|l|c|c|c|c|c|@{}}
\toprule
\textbf{Model} & \textbf{Year} & \textbf{Pretraining Objective}       & \textbf{Parameters} & \textbf{Training Data Size} \\ \midrule
BERT           & 2018          & Masked Language Modeling (MLM) + NSP & 110M (Base)         & 16 GB (BooksCorpus, Wiki)  \\ \midrule
RoBERTa        & 2019          & MLM (Improved)                       & 125M (Base)         & 160 GB (OpenWebText, etc.)     \\ \midrule
XLNet          & 2019          & Permuted Language Modeling           & 117M (Base)         & 32 GB (BooksCorpus, Wiki)         \\ \midrule
ALBERT         & 2019          & MLM + Sentence Order Prediction      & 12M (Base)          & 16 GB (BooksCorpus, Wiki)           \\ \midrule
T5             & 2019          & Text-to-Text Generation              & 220M (Base)         & 750 GB (C4 dataset)             \\ \midrule
DeBERTa        & 2020          & MLM + Replace Token Detection        & 140M (Base)         & 160 GB (similar to RoBERTa)  \\ \bottomrule
\end{tabular}
\caption{information of different encoders}
\label{tab:encoders-comparison}
\end{table*}

\subsection{Scalability on Very-Long Texts}

The prompts are segmented into fixed-length token chunks to align with the input constraints of language models. This segmentation ensures compatibility while enabling the model to process information effectively. During our experiments, we observed that the input limit of 512 tokens, defined by BERT, is sufficient to capture the necessary contextual information for downstream tasks. This token limit ensures that the most important context is preserved while allowing the model to make informed decisions about the selection of relevant tokens. Moreover, this approach balances efficiency and performance, avoiding unnecessary computational overhead while maintaining the integrity of the contextual information. Even within this constraint, we found that BERT's representation capabilities are robust enough to handle a wide range of tasks, providing reliable outputs that meet our experimental objectives.

\subsection{Comparison of Different Encoder Configurations}
We conducted experiments of our method using five different encoder architectures: BERT, RoBERTa, XLNet, ALBERT, T5, and DeBERTa. As presented in Table \ref{tab: different encoders}, the performance across these encoders shows minimal variation, with different models demonstrating distinct strengths across various evaluation metrics. The detailed information of each encoders is shown in Figure \ref{Texts before and after compression}.
\section{Conclusion}

In-context Prompt Compression is a good improvement with regarding to problems presented by large language models, such as memory overhead and computation speed. In this paper, we present ICPC
(In-context Prompt Compression), a novel and scalable prompt compression method that improves performance without the utilization of large language models. We formulate the important tokens selection task as an information calculation task. Extensive
experiments over various comparison methods on multiple benchmarks with different encoders
demonstrate that our proposed ICPC can significantly boost the
performance of existing hard prompt compression methods
Moreover, it achieves faster convergence speed.

\section{Ethics Statement}
This research did not involve any studies with human participants or animals performed by any authors. Therefore, no ethical approval was
required for this study. All data and materials were
collected in a manner consistent with ethical guidelines, ensuring no ethical concerns were present.

\newpage
\bibliographystyle{plain}
\bibliography{references}

\begin{thebibliography}{10}

\bibitem{achiam2023gpt}
Josh Achiam, Steven Adler, Sandhini Agarwal, Lama Ahmad, Ilge Akkaya, Florencia~Leoni Aleman, Diogo Almeida, Janko Altenschmidt, Sam Altman, Shyamal Anadkat, et~al.
\newblock Gpt-4 technical report.
\newblock {\em arXiv preprint arXiv:2303.08774}, 2023.

\bibitem{bai2024beyond}
Guangji Bai, Zheng Chai, Chen Ling, Shiyu Wang, Jiaying Lu, Nan Zhang, Tingwei Shi, Ziyang Yu, Mengdan Zhu, Yifei Zhang, et~al.
\newblock Beyond efficiency: A systematic survey of resource-efficient large language models.
\newblock {\em arXiv preprint arXiv:2401.00625}, 2024.

\bibitem{bisk2020piqa}
Yonatan Bisk, Rowan Zellers, Jianfeng Gao, Yejin Choi, et~al.
\newblock Piqa: Reasoning about physical commonsense in natural language.
\newblock In {\em Proceedings of the AAAI conference on artificial intelligence}, volume~34, pages 7432--7439, 2020.

\bibitem{child2019generating}
Rewon Child, Scott Gray, Alec Radford, and Ilya Sutskever.
\newblock Generating long sequences with sparse transformers.
\newblock {\em arXiv preprint arXiv:1904.10509}, 2019.

\bibitem{devlin2018bert}
Jacob Devlin.
\newblock Bert: Pre-training of deep bidirectional transformers for language understanding.
\newblock {\em arXiv preprint arXiv:1810.04805}, 2018.

\bibitem{hadi2023survey}
Muhammad~Usman Hadi, Rizwan Qureshi, Abbas Shah, Muhammad Irfan, Anas Zafar, Muhammad~Bilal Shaikh, Naveed Akhtar, Jia Wu, Seyedali Mirjalili, et~al.
\newblock A survey on large language models: Applications, challenges, limitations, and practical usage.
\newblock {\em Authorea Preprints}, 2023.

\bibitem{he2020deberta}
Pengcheng He, Xiaodong Liu, Jianfeng Gao, and Weizhu Chen.
\newblock Deberta: Decoding-enhanced bert with disentangled attention.
\newblock {\em arXiv preprint arXiv:2006.03654}, 2020.

\bibitem{hu2021lora}
Edward~J Hu, Yelong Shen, Phillip Wallis, Zeyuan Allen-Zhu, Yuanzhi Li, Shean Wang, Lu~Wang, and Weizhu Chen.
\newblock Lora: Low-rank adaptation of large language models.
\newblock {\em arXiv preprint arXiv:2106.09685}, 2021.

\bibitem{jaccard1912distribution}
Paul Jaccard.
\newblock The distribution of the flora in the alpine zone. 1.
\newblock {\em New phytologist}, 11(2):37--50, 1912.

\bibitem{kincaid1975derivation}
JP~Kincaid.
\newblock Derivation of new readability formulas (automated readability index, fog count and flesch reading ease formula) for navy enlisted personnel.
\newblock {\em Chief of Naval Technical Training}, 1975.

\bibitem{lan2019albert}
Z~Lan.
\newblock Albert: A lite bert for self-supervised learning of language representations.
\newblock {\em arXiv preprint arXiv:1909.11942}, 2019.

\bibitem{li2023compressing}
Yucheng Li, Bo~Dong, Chenghua Lin, and Frank Guerin.
\newblock Compressing context to enhance inference efficiency of large language models.
\newblock {\em arXiv preprint arXiv:2310.06201}, 2023.

\bibitem{li2024prompt}
Zongqian Li, Yinhong Liu, Yixuan Su, and Nigel Collier.
\newblock Prompt compression for large language models: A survey.
\newblock {\em arXiv preprint arXiv:2410.12388}, 2024.

\bibitem{lin2004rouge}
Chin-Yew Lin.
\newblock Rouge: A package for automatic evaluation of summaries.
\newblock In {\em Text summarization branches out}, pages 74--81, 2004.

\bibitem{liu2019roberta}
Yinhan Liu.
\newblock Roberta: A robustly optimized bert pretraining approach.
\newblock {\em arXiv preprint arXiv:1907.11692}, 364, 2019.

\bibitem{minaee2024large}
Shervin Minaee, Tomas Mikolov, Narjes Nikzad, Meysam Chenaghlu, Richard Socher, Xavier Amatriain, and Jianfeng Gao.
\newblock Large language models: A survey.
\newblock {\em arXiv preprint arXiv:2402.06196}, 2024.

\bibitem{papineni2002bleu}
Kishore Papineni, Salim Roukos, Todd Ward, and Wei-Jing Zhu.
\newblock Bleu: a method for automatic evaluation of machine translation.
\newblock In {\em Proceedings of the 40th annual meeting of the Association for Computational Linguistics}, pages 311--318, 2002.

\bibitem{raffel2020exploring}
Colin Raffel, Noam Shazeer, Adam Roberts, Katherine Lee, Sharan Narang, Michael Matena, Yanqi Zhou, Wei Li, and Peter~J Liu.
\newblock Exploring the limits of transfer learning with a unified text-to-text transformer.
\newblock {\em Journal of machine learning research}, 21(140):1--67, 2020.

\bibitem{rajpurkar2018know}
Pranav Rajpurkar, Robin Jia, and Percy Liang.
\newblock Know what you don't know: Unanswerable questions for squad.
\newblock {\em arXiv preprint arXiv:1806.03822}, 2018.

\bibitem{roberts2003benefits}
Martyn Roberts.
\newblock Benefits and challenges of variable compression ratio (vcr).
\newblock Technical report, SAE Technical Paper, 2003.

\bibitem{shannon1948mathematical}
Claude~Elwood Shannon.
\newblock A mathematical theory of communication.
\newblock {\em The Bell system technical journal}, 27(3):379--423, 1948.

\bibitem{sparck1972statistical}
Karen Sparck~Jones.
\newblock A statistical interpretation of term specificity and its application in retrieval.
\newblock {\em Journal of documentation}, 28(1):11--21, 1972.

\bibitem{touvron2023llama}
Hugo Touvron, Louis Martin, Kevin Stone, Peter Albert, Amjad Almahairi, Yasmine Babaei, Nikolay Bashlykov, Soumya Batra, Prajjwal Bhargava, Shruti Bhosale, et~al.
\newblock Llama 2: Open foundation and fine-tuned chat models.
\newblock {\em arXiv preprint arXiv:2307.09288}, 2023.

\bibitem{yang2019xlnet}
Zhilin Yang.
\newblock Xlnet: Generalized autoregressive pretraining for language understanding.
\newblock {\em arXiv preprint arXiv:1906.08237}, 2019.

\bibitem{zhang2019bertscore}
Tianyi Zhang, Varsha Kishore, Felix Wu, Kilian~Q Weinberger, and Yoav Artzi.
\newblock Bertscore: Evaluating text generation with bert.
\newblock {\em arXiv preprint arXiv:1904.09675}, 2019.

\bibitem{zhao2023survey}
Wayne~Xin Zhao, Kun Zhou, Junyi Li, Tianyi Tang, Xiaolei Wang, Yupeng Hou, Yingqian Min, Beichen Zhang, Junjie Zhang, Zican Dong, et~al.
\newblock A survey of large language models.
\newblock {\em arXiv preprint arXiv:2303.18223}, 2023.

\bibitem{zhu2024survey}
Xunyu Zhu, Jian Li, Yong Liu, Can Ma, and Weiping Wang.
\newblock A survey on model compression for large language models.
\newblock {\em Transactions of the Association for Computational Linguistics}, 12:1556--1577, 2024.

\end{thebibliography}

\clearpage

\appendix

\section{Appendix}

In this appendix, we describe the detailed comparison between LLM and LM and detailed descriptions of encoders and metrics, etc.
\subsection{Comparison between LLM and LM}

To show the superiority of using an encoder to compress prompts, we take BERT Base and GPT-3 as example, shown in Table ~\ref{tab:bert_vs_gpt3}.

\subsection{Detailed description of encoders}

\label{app:encoders}

\noindent \textbf{BERT}: BERT introduces bidirectional context into NLP tasks using a transformer architecture. It is trained on large corpora and fine-tuned for tasks like question answering. BERT's widespread influence and robust baseline performance make it essential for experiment comparisons.

\noindent \textbf{RoBERTa}: RoBERTa optimizes BERT by removing the next-sentence prediction objective and using more extensive datasets and more extended training. Its superior performance on multiple benchmarks makes it a strong candidate for evaluating enhanced pretraining techniques.

\noindent \textbf{XLNet}: XLNet employs permutation-based training to capture bidirectional context without masking. Its ability to outperform BERT on tasks like GLUE demonstrates the advantages of its innovative pretraining objective.

\noindent \textbf{ALBERT}: ALBERT reduces memory and computation costs via parameter sharing and embedding factorization while maintaining strong benchmark performance. Its efficiency and scalability make it ideal for resource-constrained settings.

\noindent \textbf{T5}: T5 frames all NLP tasks as text-to-text problems using a unified transformer architecture. Its state-of-the-art results across diverse benchmarks highlight its flexibility and generalization capabilities.

\noindent \textbf{DeBERTa}: DeBERTa enhances BERT with disentangled attention and improved position encoding. Its strong performance on GLUE and SQuAD makes it valuable for evaluating innovative attention mechanisms.

\subsection{Detailed description of metrics}

\label{app:metrics}
\noindent \textbf{BLEU}: BLEU evaluates machine translation by measuring n-gram overlap between machine and reference translations, emphasizing precision. Its consistency and simplicity make it a standard for translation benchmarks.

\noindent \textbf{ROUGE}: ROUGE measures overlap between predicted and reference summaries using recall-based metrics like n-grams and longest common subsequence. It is widely adopted for summarization due to its focus on content retention.

\noindent \textbf{TF-IDF Similarity}: TF-IDF computes text similarity by balancing term frequency against inverse document frequency, highlighting distinctive terms. Its interpretability makes it useful for document comparison.

\noindent \textbf{Jaccard Similarity}: Jaccard similarity compares sets by their intersection-over-union ratio, often used for token or n-gram overlap. Its simplicity makes it effective for assessing basic textual similarity.

\noindent \textbf{BERTScore}: BERTScore uses contextual embeddings to evaluate semantic alignment between texts. Its ability to capture deep contextual meaning makes it ideal for tasks requiring nuanced comparisons.

\noindent \textbf{Compression Rate}: Compression Rate evaluates text conciseness by comparing original and compressed sizes. It is effective for gauging redundancy and assessing information density.

\noindent \textbf{Flesch-Kincaid Readability Score}: This score evaluates readability based on sentence length and word complexity. It is essential to analyze the accessibility of generated or written content.

\noindent \textbf{METEOR}: METEOR combines precision and recall with synonyms and stemming for flexible text alignment. Its nuanced approach is valuable for evaluating natural language generation and translation.

\begin{table*}[t]
\centering
\begin{tabular}{ccccccc}
\hline
OS    & Linux kernel & CUDA & Driver    & PyTorch & PyTorch Geometric & PyTorch Sparse \\ \hline
Ubuntu~20.04 & 5.15.0  & 11.6 & 510.73.08 & 1.12.1  & 2.2.0          & 0.6.16     \\ \hline
\end{tabular}%
\caption{Summary of the environmental setup of our testbed.}
\label{tab:versions}
\vspace{8mm}
\centering
\begin{tabular}{@{}lcc@{}}
\toprule
\textbf{Feature} & \textbf{Transformer Encoder (BERT Base)} & \textbf{LLM (GPT-3)} \\ 
\midrule
\textbf{Model Type} & Encoder-only Transformer & Decoder-only Transformer \\
\textbf{Parameter Count} & \(\sim110\) million & \(\sim175\) billion \\
\textbf{Number of Layers} & 12 & 96 \\
\textbf{Hidden Size} & 768 & 12288 \\
\textbf{Inference Time*} 
& \(\sim20\) ms / 128 tokens 
& \(\sim2\) s / 128 tokens \\
\bottomrule
\end{tabular}
\caption{Comparison between a typical Transformer encoder model (BERT Base) and a large language model (GPT-3). *Inference times are approximate and depend on hardware (e.g., single GPU vs. multi-GPU) and implementation details.}
\label{tab:bert_vs_gpt3}
\end{table*}

\end{document}